\documentclass{Interspeech2024}
\usepackage{tipa}
\usepackage{xcolor}
\usepackage[english]{babel}



\interspeechcameraready

\title{Human-like Linguistic Biases in Neural Speech Models: Phonetic Categorization and Phonotactic Constraints in Wav2Vec2.0}

\name[affiliation={1}]{Marianne}{de Heer Kloots}
\name[affiliation={1}]{Willem}{Zuidema}

\address{
  $^1$Institute for Logic, Language and Computation, University of Amsterdam; The Netherlands}
\email{m.l.s.deheerkloots@uva.nl, w.h.zuidema@uva.nl}


\begin{document}

\maketitle

\begin{abstract}
    
    What do deep neural speech models know about phonology? Existing work has examined the encoding of individual linguistic units such as phonemes in these models. Here we investigate interactions between units. Inspired by classic experiments on human speech perception, we study how Wav2Vec2 resolves phonotactic constraints. We synthesize sounds on an acoustic continuum between /l/ and /r/ and embed them in controlled contexts where only /l/, only /r/, or neither occur in English. Like humans, Wav2Vec2 models show a bias towards the phonotactically admissable category in processing such ambiguous sounds. Using simple measures to analyze model internals on the level of individual stimuli, we find that this bias emerges in early layers of the model's Transformer module. This effect is amplified by ASR finetuning but also present in fully self-supervised models. Our approach demonstrates how controlled stimulus designs can help localize specific linguistic knowledge in neural speech models.

\end{abstract}

\section{Introduction}

The current state-of-the-art in speech technology largely builds on self-supervised neural architectures. These models learn representations of spoken language based only on audio, which prove to be extremely useful for downstream tasks like automatic speech recognition (ASR).
In this paper, we ask whether the latent transformations across the layers of such neural speech models also align with human speech processing steps, specifically in implicitly using knowledge about a spoken language's phonology in encoding its speech sounds. 

This is a challenging question, because these architectures are not designed as models of human processing, and exhibit major differences to human processing and learning. Both systems however share the high-level computational goal of mapping a variable acoustic signal onto generalizable discrete representations \cite{scharenborgHowShouldSpeech2005}, raising the question to what extent they may have converged to similar solutions.
Recent work shows that representations extracted from self-supervised speech models align well with linguistic annotations of phonological, lexical, and syntactic units \cite{martinProbingSelfsupervisedSpeech2023, pasadWhatSelfSupervisedSpeech2024, shenWaveSyntaxProbing2023}. 
Complementary lines of research have directly compared these representations with human neural and behavioural responses to speech stimuli \cite{liDissectingNeuralComputations2023, milletSelfsupervisedSpeechModels2022}, or investigated how the prediction errors of ASR models align with human psychometric curves under controlled stimulus distortions \cite{adolfiSuccessesCriticalFailures2023, weertsPsychometricsAutomaticSpeech2022}.

Importantly, modelling speech perception involves more than just reproducing a repertoire of responses to basic speech sounds.
Human speech perception also involves integrating low-level acoustic information with learned linguistic knowledge. 
Classic evidence for this comes from phonetic categorization experiments, in which listeners hear a range of ambiguous speech sounds on a continuum between two unambiguous phoneme categories \cite{mcqueenPhoneticCategorisation1996}. 
The categorization of such ambiguous sounds is influenced for example by the sentence context in which they are presented, and by the lexical status of the continuum endpoints \cite{connineConstraintsInteractiveProcesses1987, ganongPhoneticCategorizationAuditory1980}. 
A seminal experiment by Massaro \& Cohen demonstrated effects of phonotactic (in)admissability on phoneme categorization \cite{massaroPhonologicalContextSpeech1983}: when presented with ambiguous sounds on a continuum between /l/ and /r/ in varying consonant clusters, listeners showed a bias towards perceiving the consonant that would be phonotactically admissable in English (i.e., \textipa{/l/} when preceded by \textipa{/s/}, and \textipa{/r/} when preceded by \textipa{/t/}). 

In light of such context effects, the most promising aspect of modern speech model architectures is perhaps their ability to generate \emph{contextualized} representations of speech sounds. 
Recent evidence suggests that higher layers of Transformer-based ASR systems may abstract away from the detailed acoustic and phonetic structure encoded in earlier layers \cite{tenboschPhonemicCompetitionEndtoend2023}, and integrate information from the wider linguistic context to generate accurate orthographic transcriptions \cite{mohebbiHomophoneDisambiguationReveals2023}.
In these systems, information about phone sequence likelihoods (which in classic, modular speech recognition systems would be handled by a language modelling component) must get implicitly encoded in the model's internal weights, either during pretraining on unlabelled speech data, or from finetuning on the ASR task.

Here, we explore the possibility of using a controlled phonetic categorization experiment to examine how one prominent neural speech model -- Wav2Vec2 -- processes speech sounds, and specifically how to localize the model components in which contextual adaptation to phonotactic constraints might happen. 
In addition, we explore the presence of such phonotactic sensitivity in a range of self-supervised and ASR-finetuned Wav2Vec2 models, and the suitability of different analysis methods to study such effects. 

\begin{figure*}[ht]
  \centering
  \includegraphics[width=\textwidth]{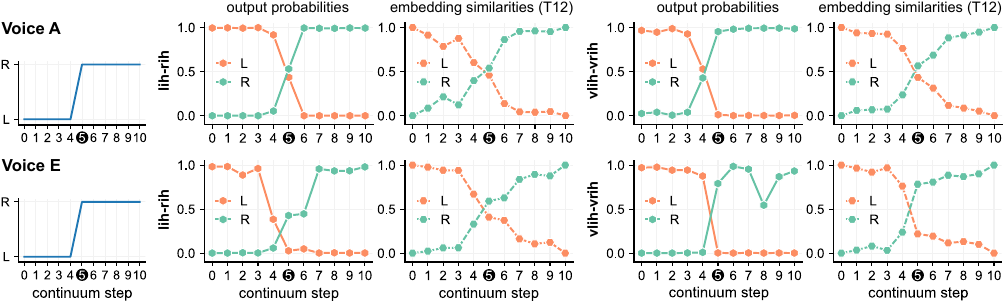}
  \caption{Responses of the ASR-finetuned Wav2Vec2-base model to the ambiguous sounds at each step of the continua between \textipa{/l/} (0) and \textipa{/r/} (10). Black circles mark the crossing points at which our response measures start indicating a preference for `R' above `L'.}
  \label{fig:probs-sims}
\end{figure*}

\section{Methods}
\subsection{Stimuli}
To investigate whether Wav2Vec2 displays consistent categorization patterns for acoustically ambiguous speech sounds, we generate continua of monosyllabic nonwords morphing between \textipa{[lI]} (`lih') and \textipa{[\*rI]} (`rih')\footnote{Massaro \& Cohen used syllables ending in \textipa{[i]}, e.g. \textipa{[tli]}-\textipa{[t\*ri]}; here we use \textipa{[I]} to avoid confounding lexical and phonotactic effects.}. We follow Massaro \& Cohen's experiment in prepending the morphing glides with consonants that allow us to examine the influence of phonotactic admissability on the model's categorization behaviour (/t/, admissable before /r/; /s/, admissable before /l/; and /v/, unlikely for either). 

For constructing the resulting nonwords of interest, we synthesized source material with American English pronunciation using two voices from the Google Text-To-Speech (TTS) API (one male, labelled \texttt{en-US-Standard-A}, and one female, labelled \texttt{en-US-Standard-E}). Directly generating phonotactically inadmissable consonant clusters (\emph{tl-}, \emph{sr-}) or words ending in \textipa{/I/} was not possible; the TTS system tends to generate character-by-character pronunciations instead. We therefore manually edited the system's generated audio for the inputs `lizz' and `rizz', removing the final sibilant to create our \emph{lih} and \emph{rih} source syllables. We then concatenated these source syllables to each of the three initial context consonants extracted respectively from the generated audio for `vizz', `tizz', and `sizz'. This procedure resulted in our 8 continuum endpoints (4 pairs: \emph{lih}-\emph{rih}, \emph{vlih}-\emph{vrih}, \emph{tlih}-\emph{trih}, \emph{slih}-\emph{srih}), in 2 voice variations each.

For morphing between these endpoints, we used an open-source GUI toolkit for the WORLD vocoder system \cite{kawaharaInteractiveToolsMaking2024, moriseWORLDVocoderBasedHighQuality2016}. WORLD is a general-purpose speech analysis and synthesis system which enables interpolation between signals through estimated parameters for fundamental frequency, spectral envelope, and aperiodic components. The morphing procedure involves several manual steps: annotating voiced and unvoiced segments, aligning the two waveforms, selecting time-frequency anchor points, and choosing the number of continuum steps. We refer to the WORLD GUI repository\footnote{\url{https://github.com/HidekiKawahara/worldGUItools}} for a full demonstration of the continuum generation pipeline, which we followed closely for each continuum. 

After generating 11-step continua between all endpoint pairs for each voice, we end up with 176 audio files which form separate input stimuli to the Wav2Vec2 model in our further analyses\footnote{All code and data needed to reproduce our analyses is available at \url{https://github.com/mdhk/phonotactic-sensitivity}}. For each continuum endpoint, we also annotated the start and end of the morphing target sound (i.e., \textipa{/l/} and \textipa{/r/}). The morphing target annotation as well as the endpoint audio composition described above were done in Praat \cite{boersmaPraatDoingPhonetics2023}.

\subsection{Wav2Vec2.0}
The Wav2Vec2 architecture \cite{baevskiWav2vecFrameworkSelfSupervised2020} consists of a 7-layer CNN module followed by a series of Transformer layers (12 in the base model, and 24 in the large version). The CNN component encodes the audio waveform into frame representations with a receptive field of 25 ms, at 20 ms intervals. These frame representations are then contextualized by the Transformer module against all other frames in the input. Wav2Vec2 is initially pretrained with a self-supervised masked frame prediction objective on 960 hours of unlabelled audio from Librispeech \cite{panayotovLibrispeechASRCorpus2015}. For ASR, the pretrained architecture is extended with a CTC \cite{gravesConnectionistTemporalClassification2006} head and finetuned with supervision to map from audio frame representation to orthographic transcriptions.
In our current study, we first consider the ASR-finetuned version of the Wav2Vec2 base model in assessing the relation between its internal representations and its output character predictions. We then compare results in base- and large-sized pretrained and finetuned models, and matching untrained architectures with randomly initialized weights. Next to the pretrained speech base model, we also include a base model pretrained on 600 hours of non-speech acoustic scenes (released by \cite{milletSelfsupervisedSpeechModels2022}).
We use the Transformers library \cite{wolfTransformersStateoftheArtNatural2020a} to download and interface with each model, as available through the HuggingFace hub.

\subsection{Model analysis methods}
For each step along our generated continua, we want to measure the model's preference for the \textipa{/l/} vs.\ \textipa{/r/} consonants. 
In the ASR-finetuned models, we can measure this preference by considering the model's output probabilities for the `L' vs.\ `R' characters, respectively. We take the maximum probability assigned to each consonant over all frames that overlap with the morphing target sound. 
Additionally, we analyze the model's internal processing of the morphing target sound using three measures that are also applicable to non-ASR-finetuned models, based on their hidden representations:

\subsubsection{Embedding similarities}
Each continuum endpoint includes an unambiguous realization of the \textipa{/l/} or \textipa{/r/} consonant. We take the model's hidden states when processing these unambiguous endpoints as reference points (`L' and `R'), and compare the states generated for the morphing sound $X$ to these references. To create vector representations $X$, `L' and `R', we average over all frame representations within the morphing target time window. We then compute our relative similarity measure based on cosine distances. For example, for the similarity ($sim$) of $X$ to `R':
\begin{align}
  sim(X, \text{`R'}) &= 1 - \frac{D_{cos}(X, \text{`R'})}{D_{cos}(X,\text{`R'})+ D_{cos}(X,\text{`L'})}
  \label{eq:sim}
\end{align}
We apply this similarity measure within layers, to the 512-dimensional output of the CNN, as well as the 768-dimensional outputs from each block in the Transformer module. 

\subsubsection{Probing classifier probabilities}
A more common way of decoding phoneme identity from model internals is by using auxiliary classifiers to predict phoneme labels based on the model's hidden layer activations \cite{alishahiEncodingPhonologyRecurrent2017c}. We trained binary logistic regression probes to predict phone labels (\emph{l} or \emph{r}) from averaged phone representations, using a set of 4000 phonetically transcribed word pronunciations from the train data of the TIMIT corpus \cite{garofolojohns.TIMITAcousticPhoneticContinuous1993}, selected for occurrences of \emph{l} and \emph{r} and balanced for speaker sex and identity. 

For all trained audio models, the classifiers reached between 90 and 99\% accuracy on a similarly balanced set of 2000 words from the TIMIT test data, with increasing accuracies for classifiers trained on later model layers. Accuracy of the classifiers trained on the untrained models ranged between 62 and 71\%. 
We then fed the frame-averaged model representations for our morphing sounds to these trained classifiers and compared the probabilities for the \emph{l} vs.\ \emph{r} labels. Similar to the embedding similarities, we extract probing classifier probabilities for the final CNN output as well as all Transformer blocks.

\subsubsection{CTC-lens probabilities}
Finally, we introduce a more experimental measure, inspired by recent theoretical and empiricial work on interpreting the internal functioning of text-based Transformers by applying the model's own unembedding operations on earlier layers \cite{nostalgebraistInterpretingGPTLogit, langedijkDecoderLensLayerwiseInterpretation2023}. For our \emph{CTC-lens}, we feed the hidden states from intermediate Transformer blocks directly into the CTC head that normally operates on the model's final layer. We then extract probabilities for the `L' and `R' characters as we would for the output layer. We also use the CTC head of the corresponding finetuned model to decode layers of the pretrained models.
Examining earlier model layers through this lens allows us to assess whether the prominence of either phoneme category found with our other methods is also structured in a way that would let the CTC head read out the corresponding orthographic character.

\section{Results}

\subsection{Locating perceptual boundaries along continua}
We start by examining responses to the ambiguous phones along our continua in the ASR-finetuned base model (\texttt{facebook/wav2vec2-base-960h}). Although we synthesized our stimuli to perceptually morph between the \textipa{/l/} and \textipa{/r/} phonemes to human ears, it does not trivially follow that Wav2Vec2's processing of these sounds would follow comparable perceptual similarity patterns. Nevertheless, in the model's output probabilities we indeed observe a high preference towards the `L' and `R' characters respectively at either continuum endpoint, and a single shift in this preference around the middle of the continuum. In \autoref{fig:probs-sims}, we visualize these output character probabilities for two continua and both voices separately, along with our embedding similarity measure at the model's output layer. We observe that the point at which the embedding similarity towards `R' overtakes the similarity towards `L' is closely aligned with the crossing point in the model's output probabilities (both occurring at continuum step 5). We note that for some continuum steps, such as steps 4-6 of the voice E \emph{lih-rih} continuum, probabilities are relatively low for either character, indicating that the model actually assigns its highest probability to another character. However, we here choose to simulate a forced choice between `R' and `L' (as in the human experiments), and therefore focus on the preferred character with the highest probability between these two options.

\subsection{Measuring sensitivity to phonological context}
\begin{figure}
  \includegraphics[width=\linewidth]{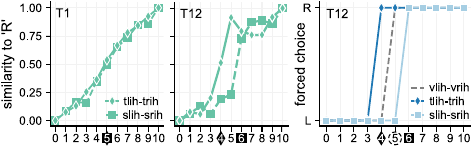}
  \caption{The model's output layer (T12) shows sensitivity to phonotactic context, based on preference for `R' using embedding similarities (Eq.\ref{eq:sim}) or max. probability forced choices.}
  \label{fig:phtc-FC}
\end{figure}

\begin{figure*}
  \centering
  \includegraphics[width=\textwidth]{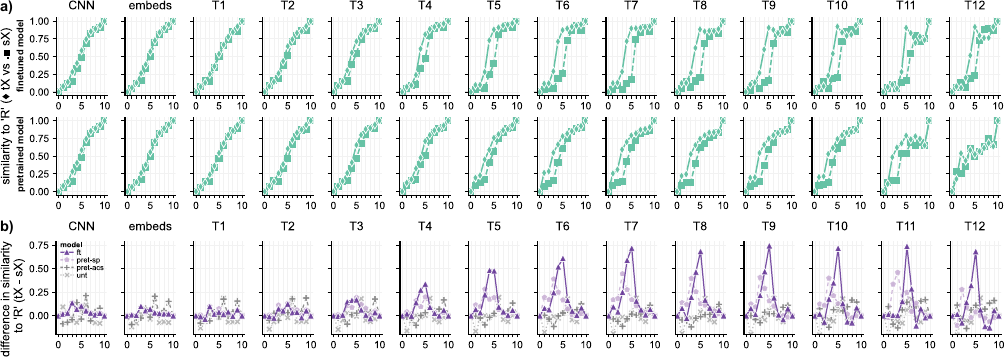}
  \caption{Phonotactic sensitivity across layers in the Wav2Vec2-base models, measured as difference in similarities to the `R' endpoint between the \emph{tXih} vs.\ \emph{sXih} continua. A bias towards the phonotactically admissable consonant can be observed in the ASR-finetuned (ft; top row), and the speech-pretrained (pret-sp; second row) model, but not in the acoustic scenes (pret-acs) and untrained (unt) model.}
  \label{fig:layers}
\end{figure*}

We proceed to compare the model's responses to the morphing target sound when a preceding consonant makes either `L' or `R' strongly preferred according to English phonotactics. Here, humans have a bias to perceive an identical target sound $X$ as `R' when preceded by t-, and as `L' when preceded by s-. \autoref{fig:phtc-FC} shows that Wav2Vec2 displays a similar bias in its preference towards the `L' vs.\ `R' consonant (we average results over the two voices from here onwards). In the \emph{tlih-trih} continuum, the switch towards preferring the `R' consonant occurs earlier (at step 4) than in the \emph{slih-srih} continuum (at step 6). This bias can also be observed as the stepwise difference in embedding similarity to `R' between the \emph{tlih-trih} and the \emph{slih-srih} continua. Using this measure, we can observe that this contextual biasing effect is present in the model's final layer (T12), but not yet in the first Transformer layer (T1). In \autoref{fig:layers}, we see that the sensitivity to phonological context starts to emerge in the finetuned model's activations around the 4th Transformer layer, and remains high until the output layer. We also find that this sensitivity tends to be highest for the most ambiguous sounds in the middle of the continuum, a pattern which generally aligns with linguistic effects on phoneme categorization in humans \cite{ganongPhoneticCategorizationAuditory1980}.

\subsection{Comparing models and analysis methods}
In \autoref{fig:layers} we also visualize the layerwise pattern of embedding similarities in the \texttt{facebook/wav2vec2-base} model which is only pretrained with self-supervision. This model shows a similar contextual bias effect emerging around the 4th or 5th Transformer layer, although less pronounced than in the ASR-finetuned model. In contrast, this effect is absent both in a randomly initialized architecture and a base model pretrained on acoustic scenes rather than speech audio, as evident by the lack of noticeable peaks for these models in \autoref{fig:layers}b.

\begin{figure}
\centering
\includegraphics[width=\linewidth]{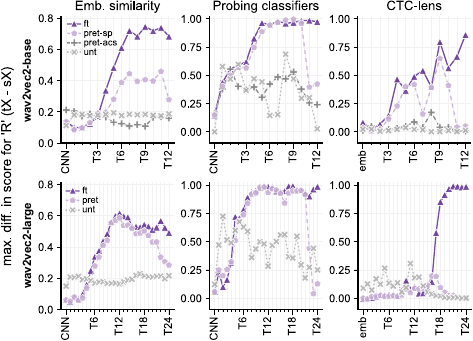}
\caption{Aggregated layerwise results for all analysis methods and two model sizes. Each point shows the highest difference in preference for `R' between the tXih and sXih continua, across all intermediate continuum steps.}
\label{fig:measure-comp}
\end{figure}

To comprehensively summarize our results for all models and analysis methods, we take the maximum difference between the observed preference for `R' in the \emph{tlih-trih} vs.\ \emph{slih-srih} conditions (i.e., the peaks in \autoref{fig:layers}b). This imperfectly summarizes the patterns discussed so far, as high peaks also occur in the noisy measurements observed for the untrained and acoustic scenes model. Nevertheless, \autoref{fig:measure-comp} reveals interesting differences between models and analysis methods. While all methods suggest some observable phonotactic bias in the pretrained base model, the embedding similarity measure still most clearly also captures apparent differences between the pretrained and finetuned representations. Moreover, when analyzing the large model architecture, the CTC-lens deviates from the other analysis measures. This could indicate that phonological information encoded in earlier layers and in the pretrained large model is only later transformed by the final layers of the finetuned model into a format that the CTC head can map to orthographic predictions.

\section{Discussion \& Conclusions}
We explored whether we can observe flexible adaptation to local phonological context in phone categorization by Wav2Vec2, by using an experimental design that closely follows classic studies on human speech perception. On continua morphing between \textipa{/l/} and \textipa{/r/}, we first found that we can locate clear crossing points where Wav2Vec2 switches from the first to the second phoneme. This suggests that Wav2Vec2 may utilize similar perceptual cues as humans in distinguishing between these two categories, even when processing highly ambiguous synthesized sounds which do not occur in its training data. 

We then set out to examine whether these crossing points also flexibly adapt when the ambiguous sounds occur in biasing phonotactic contexts. We found that the models trained on English speech replicated a human-like bias towards categorizing the ambiguous sound as the phone that is more likely to follow the preceding consonant in English. With a simple measure based on distances in the model's embedding space, we can observe this bias in the model with supervised finetuning on the ASR task, but also in the fully self-supervised pretrained model. 

This indicates that a symbolic training objective like character prediction is not necessary for the Wav2Vec2 model to implicitly learn information about English phonotactic structure. Self-supervised training on speech data with the masked frame prediction objective could be enough for the contextualization module to learn to integrate the likelihood of phones given neighbouring sounds. However, we note that our exploratory experiments are not enough to draw general conclusions on the prevalence of phonotactic sensitivity in self-supervised models. More experiments studying a greater variety of phonological contexts are needed to characterize to what extent such models apply language-specific constraints on phone combinations in creating contextualized representations of speech sounds. 

More broadly, our study demonstrates how controlled experimental paradigms from studies on human speech perception can be fruitfully applied to localize linguistic contextualization in neural speech models. This opens interesting possibilities for future work, as similar phonetic categorization studies could examine the presence of more abstract (e.g., lexical and syntactic) biases, and their robustness across different model architectures. We also compared three methods for analyzing the model's internals to explore their potential for distinctively revealing such linguistic biases. Different methods may yield complementary results \cite{chrupalaAnalyzingAnalyticalMethods2020a}, but a simple measure based on cosine distances between layer activations proved insightful for our experiments, while also staying faithful to distinct characteristics of different models' representational spaces. Overall, we hope to have demonstrated how combining carefully controlled stimuli with model interpretability techniques can lead to a better mechanistic understanding of spoken language processing in neural speech models.

\section{Acknowledgements}
MdHK is funded by the Netherlands Organization for Scientific Research (NWO), through a Gravitation Grant 024.001.006 to the Language in Interaction Consortium.

\bibliographystyle{IEEEtran}
\bibliography{main}

\end{document}